 \documentclass[letterpaper, 10 pt, conference]{ieeeconf}                                                          % if you need a4paper
\IEEEoverridecommandlockouts                              % This command is only
\usepackage{graphics} % for pdf, bitmapped graphics files
\usepackage{epsfig} % for postscript graphics files
\usepackage{amsmath} % assumes amsmath package installed
\usepackage{amssymb}  % assumes amsmath package installed
  \usepackage{amsthm}
\usepackage{algorithm}
\usepackage{algpseudocode}
\usepackage{hyperref}

\title{\LARGE \bf
    Episodic Learning with Control Lyapunov Functions for \\ Uncertain Robotic Systems*
}

\author{Andrew J. Taylor$^{1}$, Victor D. Dorobantu$^{1}$, Hoang M. Le, Yisong Yue, Aaron D. Ames% <-this % stops a space
\thanks{*This work was supported by Google Brain Robotics and DARPA Award HR00111890035}% <-this % stops a space
\thanks{$^{1}$Both authors contributed equally.}% 
\thanks{All authors are with the Department of Computing and Mathematical Sciences, California Institute of Technology, Pasadena, CA 91125, USA {\tt\small ajtaylor@caltech.edu, vdoroban@caltech.edu, hmle@caltech.edu, yyue@caltech.edu, ames@caltech.edu}}%
}

\newcommand{\der}[2]{\frac{\mathrm{d} #1 }{\mathrm{d} #2 }}
\newcommand{\derp}[2]{\frac{\partial #1 }{\partial #2 }}
\newcommand{\norm}[1]{\left\Vert #1 \right\Vert}

\renewcommand{\cal}[1]{\mathcal{ #1 }}
\newcommand{\mb}[1]{\mathbf{ #1 }}
\newcommand{\bs}[1]{\boldsymbol{ #1 }}

\newcommand{\R}{\mathbb{R}}

\newcommand{\union}{\cup}

\DeclareMathOperator*{\argmin}{arg\,min}

\newtheorem{assumption}{Assumption}

\begin{document}

\maketitle
\thispagestyle{empty}
\pagestyle{empty}

%%%%%%%%%%%%%%%%%%%%%%%%%%%%%%%%%%%%%%%%%%%%%%%%%%%%%%%%%%%%%%%%%%%%%%%%%%%%%%%%
\begin{abstract}
Many modern nonlinear control methods aim to endow systems with guaranteed properties, such as stability or safety, and have been successfully applied to the domain of robotics. However, model uncertainty remains a persistent challenge, weakening theoretical guarantees and causing implementation failures on physical systems. This paper develops a machine learning framework centered around Control Lyapunov Functions (CLFs) to adapt to parametric uncertainty and unmodeled dynamics in general robotic systems. Our proposed method proceeds by iteratively updating estimates of Lyapunov function derivatives and improving controllers, ultimately yielding a stabilizing quadratic program model-based controller. We validate our approach on a planar Segway simulation, demonstrating substantial performance improvements by iteratively refining on a base model-free controller.
\end{abstract}
%%%%%%%%%%%%%%%%%%%%%%%%%%%%%%%%%%%%%%%%%%%%%%%%%%%%%%%%%%%%%%%%%%%%%%%%%%%%%%%%

\section{INTRODUCTION}
The use of Control Lyapunov Functions (CLFs) \cite{artstein1983stabilization,Sontag:universal} for nonlinear control of robotic systems is becoming increasingly popular \cite{ma2017bipedal,galloway2015torque,nguyen2015optimal}, often utilizing quadratic program controllers \cite{ames2013towards,ames2014rapidly,galloway2015torque}.  While effective, one major challenge is the need for extensive tuning, which is largely due to modeling deficiencies such as parametric error and unmodeled dynamics (cf. \cite{ma2017bipedal}).  While there has been much research in developing robust control methods that maintain stability under  uncertainty (e.g., via input-to-state stability \cite{sontag2008input}) or in adapting to limited forms of uncertainty (e.g., adaptive control \cite{krstic1995control},\cite{karafyllis2018adaptive}), relatively little work has been done on systematically reducing uncertainty while maintaining stability for general function classes of models.

%The method of nonlinear control for robotic systems using Control Lyapunov Functions (CLFs), originally developed in \cite{artstein1983stabilization}, \cite{Sontag:universal}, has recently been revitalized by the discovery of its connections with quadratic programs \cite{ames2013towards}, \cite{ames2014rapidly}. These methods have been demonstrated successfully on robotic platforms but require extensive tuning upon initial deployment \cite{ma2017bipedal}. This tuning is due to the methods being model-based, making them susceptible to uncertainty through parametric error and unmodeled dynamics. The impact of this uncertainty on the performance of the system is typically considered through the lenses of robustness and adaptation. Input-to-state stability, developed in \cite{sontag1995characterizations}, \cite{sontag2008input}, can be used to describe how the stability and performance of a system degrades with the magnitude of the uncertainty present in the model \cite{kolathaya2017parameter}. While this provides a means to assess the robustness of the system, it does not allow one to reduce the impact of uncertainty on the performance. Adaptive control methods provide a means of mitigating the impact of uncertainty by updating estimates of model parameters to ensure stability \cite{krstic1995control}. These methods do not necessarily learn the correct parameter values, and many require that a system be written in an affine regressor form \cite{lu1993regressor}. This limits the scope of unmodeled dynamics that can be captured.

We take a machine learning approach to address the above limitations. Learning-based approaches have already shown great promise for controlling imperfectly modeled robotic platforms \cite{kober2013reinforcement,schaal2010learning}.
%The introduction of machine learning to the field of robotics has presented many new methods for controlling imperfectly modeled robotic platforms \cite{kober2013reinforcement}, \cite{schaal2010learning}. 
Successful learning-based approaches have typically focused on
learning model-based uncertainty \cite{beckers2019stable,berkenkamp2016safe,berkenkamp2015safe,shi2019neural}, or direct model-free controller design \cite{lillicrap2015continuous,schulman2017proximal,duan2016benchmarking,tan2018sim,le2016smooth}.

We are particularly interested in learning-based approaches that guarantee Lyapunov stability \cite{Khalil}.
%One particular branch of this work has focused on the use of Lyapunov stability from nonlinear systems theory \cite{Khalil}.
From that perspective, the bulk of previous work has focused on using learning to construct a Lyapunov function \cite{richards2018lyapunov,chow2018lyapunov,ravanbakhsh2017learning}, or to assess the region of attraction for a Lyapunov function \cite{berkenkamp2017safe,berkenkamp2016safe2}. One limitation of previous work is the learning is conducted over the full-dimensional state space, which can be data inefficient.  We instead constructively prescribe a CLF, and focus on learning only the necessary information to choose control inputs that achieve the associated stability guarantees, which can be much lower-dimensional.

One challenge in developing learning-based methods for controller improvement is how best to collect training data that accurately reflects the desired operating environment and control goals. 
In particular, exhaustive data collection typically scales exponentially with  dimensionality of the joint state and control output space, and so should be avoided.  But first pre-collecting data upfront can lead to poor performance as downstream control behavior may enter states that are not present in the pre-collected training data.  We will leverage episodic learning approaches such as Dataset Aggregation (DAgger) \cite{Ross2010DAgger} to address these challenges in a data-efficient manner, and lead to iteratively refined controllers.

%Gathering data in this environment may require iterating upon a baseline control algorithm until the system can sufficiently explore the region of interest. In addition, data collected from experiments on robotic platforms are neither independently or identically distributed, violating a fundamental assumption in analysis of supervised learning problems. Methods such as Dataset Aggregation (DAgger) \cite{Ross2010DAgger} address these problems through episodic data collection and improvement of policies.

\begin{figure}[t]
    \centering
    \begin{minipage}{0.1\textwidth}
        \hspace{0.8 in}
        \includegraphics[scale=0.18, clip, trim={9in, 0in, 9in, 1in}]{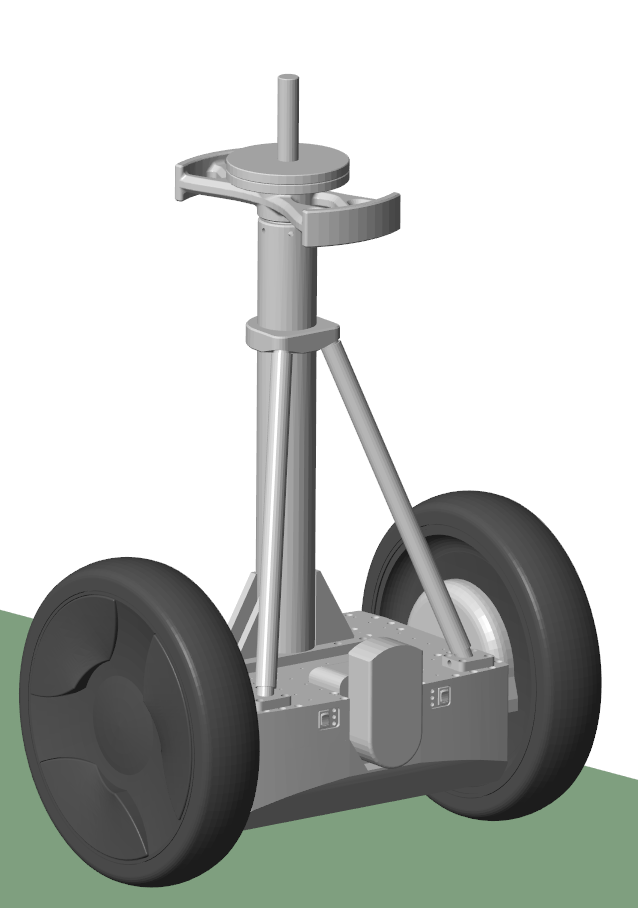}
    \end{minipage}
    \begin{minipage}{0.1\textwidth}
        \hspace{0.1 in}
        \includegraphics[scale=0.125, clip, trim={8.5in, 0.1in, 7.5in, 1in}]{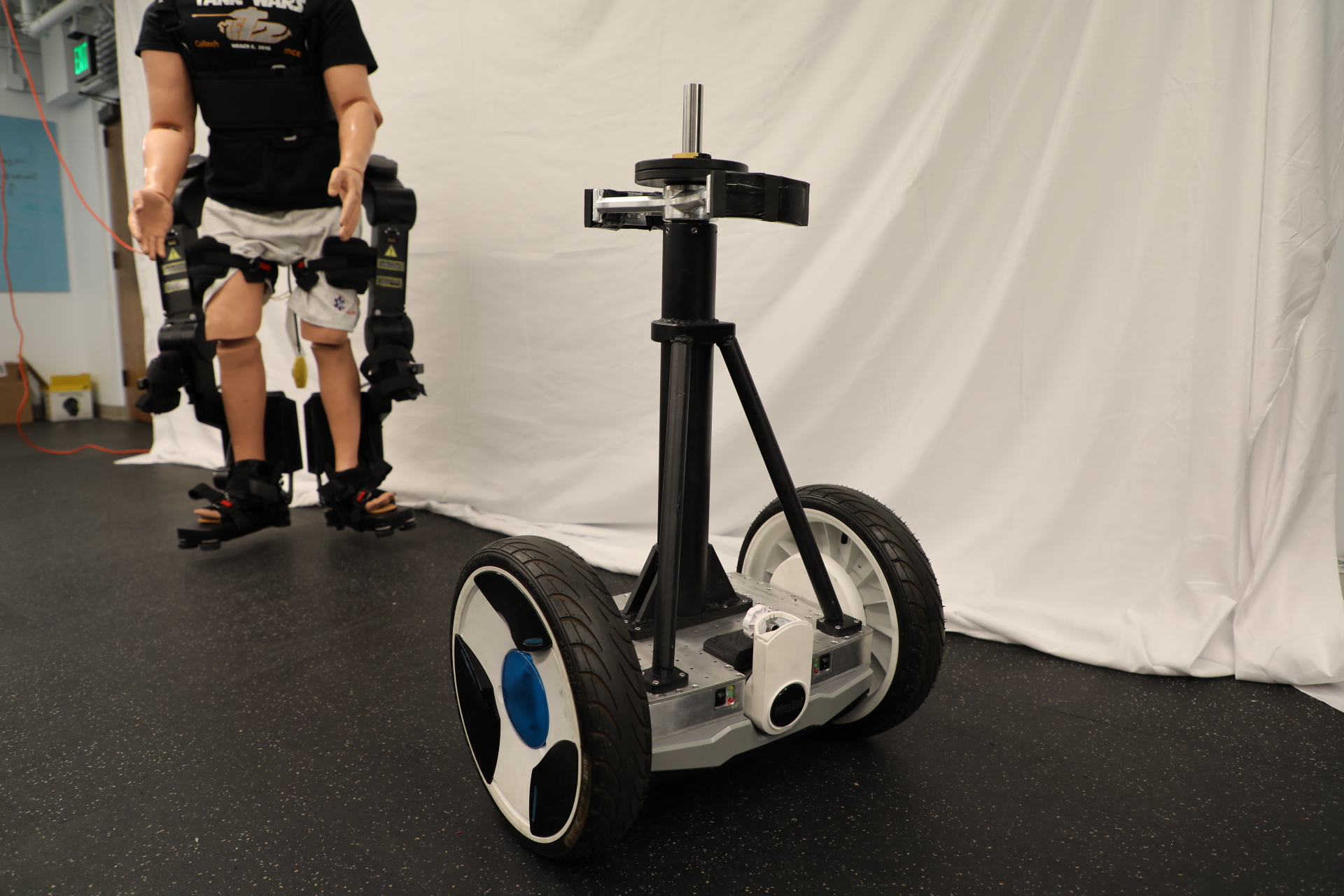}
    \end{minipage}
    \caption{CAD model \& physical system, a modified Ninebot Segway.}
    \label{fig:segway}
\end{figure}

In this paper we present a novel episodic learning approach that utilizes CLFs to iteratively improve controller design while maintaining stability. 
To the best of our knowledge, our approach is the first that integrates CLFs and general supervised learning (e.g., including deep learning) in a mathematically integrated way.  Another distinctive aspect is that our approach performs learning on the projection of state dynamics onto the CLF time derivative, which can be much lower dimensional than learning the full state dynamics or the region of attraction.
%Our approach can also naturally utilize powerful modern learning techniques such as deep learning.

Our paper is organized as follows.
Section \ref{sec:prelims} presents a review of input-output feedback linearization with a focus on constructing CLFs for unconstrained robotic systems. Section \ref{sec:uncertainty} discusses model uncertainty of a general robotic system and establishes assumptions on the structure of this uncertainty. These assumptions allow us to prescribe a CLF for the true system, but leave open the question of how to model its time derivative. Section \ref{sec:learnclfs} provides an episodic learning approach to iteratively improving a model of the time derivative of the CLF. We also present a variant of optimal CLF-based control that integrates the learned representation. Finally, Section \ref{sec:simulation} provides simulation results on a model of a modified Ninebot by Segway E+, seen in Fig. \ref{fig:segway}. We also provide a Python software package (LyaPy) implementing our experiments and learning framework.\footnote{\href{https://github.com/vdorobantu/lyapy}{https://github.com/vdorobantu/lyapy}}

\section{PRELIMINARIES ON CLFs}
\label{sec:prelims}
This section provides a brief review of input-output feedback linearization, a control technique which can be used to synthesize a CLF. The resulting CLF will be used to quantify the impact of model uncertainty and specify the learning problem outlined in Section \ref{sec:uncertainty}.

\subsection{Input-Output Linearization}
Input-Output Linearization is a nonlinear control method that creates stable linear dynamics for a selected set of outputs of a system \cite{Khalil}. The relevance of Input-Output Linearization is that it provides a constructive way to generate Lyapunov functions for the class of affine robotic control systems. Consider a configuration space $\mathcal{Q} \subseteq \mathbb{R}^n$ and an input space $\mathcal{U} \subseteq \mathbb{R}^m$. Assume $\mathcal{Q}$ is path-connected and non-empty. Consider a control system specified by:
\begin{equation}
\label{eqn:nomdyn}
    \mb{D}(\mb{q})\ddot{\mb{q}} + \underbrace{\mb{C}(\mb{q}, \dot{\mb{q}})\dot{\mb{q}} + \mb{G}(\mb{q})}_{\mb{H}(\mb{q}, \dot{\mb{q}})} = \mb{B}\mb{u},
\end{equation}
with generalized coordinates $\mb{q} \in \mathcal{Q}$, coordinate rates $\dot{\mb{q}} \in \R^n$, input $\mb{u} \in \mathcal{U}$, inertia matrix $\mb{D}: \cal{Q} \to \mathbb{S}^n_{++}$, centrifugal and Coriolis terms $\mb{C}: \cal{Q} \times \R^n \to \R^{n \times n}$, gravitational forces $\mb{G}: \cal{Q} \to \R^n$, and static actuation matrix $\mb{B} \in \R^{n \times m}$. Here $\mathbb{S}^n_{++}$ denotes the set of $n\times n$ symmetric positive definite matrices. Define twice-differentiable outputs $\mb{y} : \mathcal{Q} \rightarrow \mathbb{R}^k$, with $k \leq m$, and assume each output has relative degree 2 on some domain $\cal{R} \subseteq \cal{Q}$ (see \cite{Sastry99} for more details). Consider the time interval $\mathcal{I} = [t_0, t_f]$ for initial and final times $t_0$, $t_f$ satisfying $t_f > t_0$ and define twice-differentiable time-dependent desired outputs $\mb{y}_d : \mathcal{I} \rightarrow \mathbb{R}^k$ with $\mb{r}(t) = \begin{bmatrix}\mb{y}_d(t)^\top & \dot{\mb{y}}_d(t)^\top \end{bmatrix}^\top$. The error between the outputs and the desired outputs (commonly referred to as virtual constraints \cite{WEetal07a}) yields the dynamic system:
\begin{align}\label{eqn:fulldynamics}
    \der{}{t} \begin{bmatrix} \mb{y}(\mb{q}) - \mb{y}_d(t) \\ \dot{\mb{y}}(\mb{q}, \dot{\mb{q}}) - \dot{\mb{y}}_d(t) \end{bmatrix} &= \overbrace{\begin{bmatrix} \derp{\mb{y}}{\mb{q}}\dot{\mb{q}} \\ \derp{\dot{\mb{y}}}{\mb{q}}\dot{\mb{q}}  - \derp{\mb{y}}{\mb{q}}\mb{D}(\mb{q})^{-1}\mb{H}(\mb{q}, \dot{\mb{q}}) \end{bmatrix}}^{\mb{f}(\mb{q},\dot{\mb{q}})} \nonumber \\ &\quad\quad - \underbrace{\vphantom{\begin{bmatrix} \derp{\mb{y}}{\mb{q}}\dot{\mb{q}} \\ \derp{\dot{\mb{y}}}{\mb{q}}\dot{\mb{q}} - \derp{\mb{y}}{\mb{q}}\mb{D}(\mb{q})^{-1}\mb{H}(\mb{q}, \dot{\mb{q}}) \end{bmatrix}}\begin{bmatrix} \dot{\mb{y}}_d(t) \\ \ddot{\mb{y}}_d(t) \end{bmatrix}}_{\dot{\mb{r}}(t)} + \underbrace{\vphantom{\begin{bmatrix} \derp{\mb{y}}{\mb{q}}\dot{\mb{q}} \\ \derp{\dot{\mb{y}}}{\mb{q}}\dot{\mb{q}} - \derp{\mb{y}}{\mb{q}}\mb{D}(\mb{q})^{-1}\mb{H}(\mb{q}, \dot{\mb{q}}) \end{bmatrix}}\begin{bmatrix} \mb{0}_{k \times m} \\ \derp{\mb{y}}{\mb{q}}\mb{D}(\mb{q})^{-1}\mb{B} \end{bmatrix}}_{\mb{g}(\mb{q})}\mb{u},
\end{align}
noting that $\derp{\dot{\mb{y}}}{\dot{\mb{q}}} = \derp{\mb{y}}{\mb{q}}$. For all $\mb{q} \in \cal{R}$, $\mb{g}(\mb{q})$ is full rank by the relative degree assumption. Define $\bs{\eta} : \mathcal{Q}\times\mathbb{R}^n \times\mathcal{I} \rightarrow \mathbb{R}^{2k}$, $\widetilde{\mb{f}}: \cal{Q} \times \R^n \to \R^{k}$, and $\widetilde{\mb{g}}: \cal{Q} \to \R^{k \times m}$ as:
\begin{align}
    \bs{\eta}(\mb{q},\dot{\mb{q}}, t) &= \begin{bmatrix}\mb{y}(\mb{q}) - \mb{y}_{d}(t) \\  \dot{\mb{y}}(\mb{q}, \dot{\mb{q}}) - \dot{\mb{y}}_d(t) \end{bmatrix}\\
    \widetilde{\mb{f}}(\mb{q}, \dot{\mb{q}}) &= \derp{\dot{\mb{y}}}{\mb{q}}\dot{\mb{q}} - \derp{\mb{y}}{\mb{q}}\mb{D}(\mb{q})^{-1}\mb{H}(\mb{q}, \dot{\mb{q}})\\
    \widetilde{\mb{g}}(\mb{q}) &= \derp{\mb{y}}{\mb{q}}\mb{D}(\mb{q})^{-1}\mb{B},
\end{align}
and assume $\mathcal{U} = \mathbb{R}^m$. The input-output linearizing control input is specified by:
\begin{equation}
\label{eqn:fblctrllaw}
    \mb{u}(\mb{q}, \dot{\mb{q}}, t) = \widetilde{\mb{g}}(\mb{q})^\dagger(-\widetilde{\mb{f}}(\mb{q}, \dot{\mb{q}}) + \ddot{\mb{y}}_d(t) + \bs{\nu}(\mb{q}, \dot{\mb{q}}, t)),
\end{equation}
with auxiliary input $\bs{\nu}(\mb{q}, \dot{\mb{q}}, t)\in\mathbb{R}^k$ for all $\mb{q} \in \cal{Q}$, $\dot{\mb{q}} \in \R^n$, and $t \in \cal{I}$, where $\dagger$ denotes the Moore-Penrose pseudoinverse. This controller used in (\ref{eqn:fulldynamics}) generates linear output dynamics of the form:
\begin{equation}\label{eqn:outputaux}
    \dot{\bs{\eta}}(\mb{q}, \dot{\mb{q}}, t) = \underbrace{\begin{bmatrix} \mb{0}_{k \times k} & \mb{I}_{k \times k} \\ \mb{0}_{k \times k} & \mb{0}_{k \times k} \end{bmatrix}}_{\mb{F}} \bs{\eta}(\mb{q}, \dot{\mb{q}}, t) + \underbrace{\begin{bmatrix} \mb{0}_{k \times k}  \\ \mb{I}_{k \times k}\end{bmatrix}}_{\mb{G}}\bs{\nu}(\mb{q}, \dot{\mb{q}}, t),
\end{equation}
where $(\mb{F}, \mb{G})$ are a controllable pair. Defining $\mb{K} = \begin{bmatrix}\mb{K}_p & \mb{K}_d \end{bmatrix}$ where $\mb{K}_p,\,\mb{K}_d\in\mathbb{S}^k_{++}$, the auxiliary control input $\bs{\nu}(\mb{q}, \dot{\mb{q}},t)=-\mb{K}\bs{\eta}(\mb{q}, \dot{\mb{q}},t)$ induces output dynamics:
\begin{equation}
    \dot{\bs{\eta}}(\mb{q}, \dot{\mb{q}},t) = \mb{A}_{cl}\bs{\eta}(\mb{q}, \dot{\mb{q}},t),
    \label{eqn:outputlindyn}
\end{equation}
where $\mb{A}_{cl} = \mb{F} - \mb{G}\mb{K}$ is Hurwitz. This implies the desired output trajectory $\mb{y}_d$ is exponentially stable. This conclusion allows us to construct a Lyapunov function for the system using converse theorems found in \cite{Khalil}. With $\mb{A}_{cl}$ Hurwitz, for any $\mb{Q}\in\mathbb{S}_{++}^{2k}$, there exists a unique $\mb{P}\in\mathbb{S}_{++}^{2k}$ such that the Continuous Time Lyapunov Equation (CTLE):
\begin{equation}
    \mb{A}_{cl}^\top\mb{P} + \mb{P}\mb{A}_{cl} = -\mb{Q},
\end{equation}
is satisfied. Let $\mathcal{C} = \{ \bs{\eta}(\mb{q}, \dot{\mb{q}}, t) : (\mb{q}, \dot{\mb{q}}) \in \mathcal{R}\times\mathbb{R}^n, t \in \mathcal{I} \}$. Then $V(\bs{\eta}) = \bs{\eta}^\top\mb{P}\bs{\eta}$, implicitly a function of $\mb{q}$, $\dot{\mb{q}}$, and $t$, is a Lyapunov function certifying exponential stability of (\ref{eqn:outputlindyn}) on $\cal{C}$ satisfying:
\begin{align}\label{eqn:explyap}
    \lambda_{\mathrm{min}}(\mb{P}) \Vert \bs{\eta} \Vert^2_2 \leq V(\bs{\eta}) \leq \lambda_{\mathrm{max}}(\mb{P}) \Vert \bs{\eta} \Vert^2_2\nonumber\\
    \dot{V}(\bs{\eta}) \leq -\lambda_{\mathrm{min}}(\mb{Q}) \norm{\bs{\eta}}_2^2,
\end{align}
for all $\bs{\eta} \in \mathcal{C}$. Here $\lambda_{\mathrm{min}}(\cdot)$ and $\lambda_{\mathrm{max}}(\cdot)$ denote the minimum and maximum eigenvalues of a symmetric matrix, respectively. Alternatively, a Lyapunov function of the same form can be constructed directly from (\ref{eqn:outputaux}) using the Continuous Algebraic Riccati Equation (CARE) \cite{Khalil}.
% \begin{equation}
%     \label{eqn:algricatti}
%     \mb{F}^\top\mb{P} + \mb{P}\mb{F} - \mb{P}\mb{G}\mb{G}^\top\mb{P} = -\mb{Q},
% \end{equation}
% for any $\mb{Q} \in \mathbb{S}^{2k}_{++}$. The resulting Lyapunov function and pair $(\mb{P}, \mb{Q})$ also satisfy (\ref{eqn:explyap}) for some choice of state-feedback controller.

\subsection{Control Lyapunov Functions}
The preceding formulation of a Lyapunov function required the choice of the specific control law given in (\ref{eqn:fblctrllaw}). For optimality purposes, it may be desirable to choose a different control input for the system, thus motivating the following definition. Let $\mathcal{C} \subseteq \mathbb{R}^{2k}$. A function $V : \mathbb{R}^{2k} \rightarrow \mathbb{R}_+$ is a Control Lyapunov Function (CLF) for (\ref{eqn:nomdyn}) on $\cal{C}$ certifying exponential stability if there exist constants $c_1, c_2, c_3 > 0$ such that:
\begin{align}
    \label{eqn:clfcond}
    c_1 \Vert \bs{\eta} \Vert_2^2 \leq V(\bs{\eta}) \leq c_2 \Vert \bs{\eta} \Vert_2^2 \nonumber\\
    \inf_{\mb{u} \in \cal{U}} \dot{V}(\bs{\eta}, \mb{u}) \leq -c_3 \Vert\bs{\eta}\Vert_2^2,
\end{align}
for all $\bs{\eta} \in \mathcal{C}$. We see that the previously constructed Lyapunov function satisfying (\ref{eqn:explyap}) satisfies (\ref{eqn:clfcond}) by choosing the control input specified in (\ref{eqn:fblctrllaw}). In the absence of a specific control input, we may write the Lyapunov function time derivative as:
\begin{equation}
\dot{V}(\bs{\eta},\mb{u}) = \derp{V}{\bs{\eta}}\dot{\bs{\eta}} = \derp{V}{\bs{\eta}}(\mb{f}(\mb{q},\dot{\mb{q}})-\dot{\mb{r}}(t)+\mb{g}(\mb{q})\mb{u}).  
\end{equation}
Information about the dynamics is encoded within the scalar function $\dot{V}$, offering a reduction in dimensionality which will become relevant later in learning. Also note that $\dot{V}$ is affine in $\mb{u}$. This leads to the class of quadratic program based controllers given by:

\begin{align}
\label{eqn:qpcontroller}
\mb{u}(\mb{q},\dot{\mb{q}},t) = \argmin_{\mb{u}\in\cal{U}} ~&~ \frac{1}{2}\mb{u}^\top\mb{Mu}+\mb{s}^\top\mb{u}+r \nonumber \\\mathrm{s.t. }~&~\dot{V}(\bs{\eta},\mb{u})\leq -c_3\norm{\bs{\eta}}^2_2,
\end{align}
for $\mb{M} \in \mathbb{S}^m_+$, $\mb{s} \in \R^m$, and $r \in \R$, provided $\cal{U}$ is a polyhedron. Here $\mathbb{S}^m_+$ denotes the set of $m \times m$ symmetric positive semi-definite matrices.

\section{UNCERTAINTY MODELS \& LEARNING}
\label{sec:uncertainty}
This section defines the class of model uncertainty we consider in this work and investigates its impact on the control system, and concludes with motivation for a data-driven approach to mitigate this impact. 

\subsection{Uncertainty Modeling Assumptions}
As defined in Section \ref{sec:prelims}, we consider affine robotic control systems that evolve under dynamics described by (\ref{eqn:nomdyn}). In practice, we do not know the dynamics of the system exactly, and instead develop our control systems using the estimated model:
\begin{equation}
\label{eqn:estdyn}
    \widehat{\mb{D}}(\mb{q})\ddot{\mb{q}} + \underbrace{\widehat{\mb{C}}(\mb{q}, \dot{\mb{q}})\dot{\mb{q}} + \widehat{\mb{G}}(\mb{q})}_{\widehat{\mb{H}}(\mb{q}, \dot{\mb{q}})} = \widehat{\mb{B}}\mb{u}.
\end{equation}
We assume the estimated model \eqref{eqn:estdyn} satisfies the relative degree condition on  the domain $\cal{R}$, and thus may use the method of feedback linearization to produce a Control Lyapunov Function (CLF), $V$, for the system. Using the definitions established in (\ref{eqn:fulldynamics}) in conjunction with the estimated model, we see that true system evolves as:
\begin{align}
    \label{eqn:distaffsys}
    \dot{\bs{\eta}} = ~&~ \widehat{\mb{f}}(\mb{q},\dot{\mb{q}}) - \dot{\mb{r}}(t) + \widehat{\mb{g}}(\mb{q})\mb{u} \nonumber \\ ~&~ + (\underbrace{\mb{g}(\mb{q}) - \widehat{\mb{g}}(\mb{q})}_{\mb{A}(\mb{q})})\mb{u} + \underbrace{\mb{f}(\mb{q},\dot{\mb{q}}) - \widehat{\mb{f}}(\mb{q},\dot{\mb{q}})}_{\mb{b}(\mb{q},\dot{\mb{q}})}.
\end{align}
We note the following features of modeling uncertainty in this fashion:
\begin{itemize}
    \item Uncertainty is allowed to enter the system dynamics via parametric error as well as through completely unmodeled dynamics. In particular, the function $\mb{H}$ can capture a wide variety of nonlinear behavior and only needs to be Lipschitz continuous.
    \item This formulation explicitly allows uncertainty in how the input is introduced into the dynamics via uncertainty in the inertia matrix $\mb{D}$ and static actuation matrix $\mb{B}$. This definition of uncertainty is also compatible with a dynamic actuation matrix $\mb{B}:\cal{Q}\times\mathbb{R}^n\to\R^{n\times m}$ given proper assumptions on the relative degree of the system.
\end{itemize}
Given this formulation of our uncertainty, we make the following assumptions of the true dynamics:
\begin{assumption}
The true system is assumed to be deterministic, time invariant, and affine in the control input.
\end{assumption}
\begin{assumption}\label{as:clftruesys}
The CLF $V$, formulated for the estimated model, is a CLF for the true system.
\end{assumption}
It is sufficient to assume that the true system have relative degree 2 on the domain $\cal{R}$ to satisfy Assumption \ref{as:clftruesys}. This holds since the true values of $\widetilde{\mb{f}}$ and $\widetilde{\mb{g}}$, if known, enable choosing control inputs as in (\ref{eqn:fblctrllaw}) that respect the same linear output dynamics (\ref{eqn:outputlindyn}). Given that $V$ is a CLF for the true system, its time derivative under uncertainty is given by:
\begin{align}
\label{eq:lyapder}
    \dot{V}(\bs{\eta}, \mb{u}) = ~&~ \overbrace{\derp{V}{\bs{\eta}}( \widehat{\mb{f}}(\mb{q},\dot{\mb{q}}) - \dot{\mb{r}}(t) + \widehat{\mb{g}}(\mb{q})\mb{u})}^{\widehat{\dot{V}}(\bs{\eta}, \mb{u})} \nonumber \\ ~&~ + \underbrace{\derp{V}{\bs{\eta}}\mb{A}(\mb{q})}_{\mb{a}(\bs{\eta},\mb{q})^\top}\mb{u} + \underbrace{\derp{V}{\bs{\eta}}\mb{b}(\mb{q},\dot{\mb{q}})}_{b(\bs{\eta},\mb{q},\dot{\mb{q}})},
\end{align}
for all $\bs{\eta} \in \R^{2k}$ and $\mb{u} \in \cal{U}$. While $V$ is a CLF for the true system, it is no longer possible to determine if a specific control value will satisfy the derivative condition in (\ref{eqn:clfcond}) due to the unknown components $\mb{a}$ and $b$. Rather than form a new Lyapunov function, we seek to better estimate the Lyapunov function derivative $\dot{V}$ to enable control selection that satisfies the exponential stability requirement. This estimate should be affine in the control input, enabling its use in the controller described in (\ref{eqn:qpcontroller}). Instead of learning the unknown dynamics terms $\mb{A}$ and $\mb{b}$, which scale with both the dimension of the configuration space and the number of inputs, we will learn the terms $\mb{a}$ and $b$, which scale only with the number of inputs. In the case of the planar Segway model we simulate, we reduce the number of learned components from 4 to 2 (assuming kinematics are known). These learned representations need to accurately capture the uncertainty over the domain in which the system is desired to evolve to ensure stability during operation.

\subsection{Motivating a Data-Driven Learning Approach}
\label{sec:uncertainty_learning}
The formulation from (\ref{eqn:distaffsys}) and (\ref{eq:lyapder}) defines a general class of dynamics uncertainty. It is natural to consider a data-driven method to estimate the unknown quantities $\mb{a}$ and $b$ over the domain of the system. To motivate our learning-based framework, first consider a simple approach of learning $\mb{a}$ and $b$ via supervised regression \cite{gyorfi2006distribution}: we operate the system using some given state-feedback controller to gather data points along the system's evolution and learn a function that approximates $\mb{a}$ and $b$ via supervised learning.

Concretely, let $\mb{q}_0 \in \cal{Q}$ be an initial configuration. An experiment is defined as the evolution of the system over a finite time interval from the initial condition $(\mb{q}_0, \mb{0})$ using a discrete-time implementation of the given controller. A resulting discrete-time state history is obtained, which is then transformed with Lyapunov function $V$ and finally differentiated numerically to estimate $\dot{V}$ throughout the experiment. This yields a data set comprised of input-output pairs:
\begin{equation}
    D = \{((\mb{q}_i, \dot{\mb{q}}_i, \bs{\eta}_i, \mb{u}_i), \dot{V}_i)\}_{i=1}^N \subseteq (\cal{Q} \times \R^n \times \R^{2k} \times \cal{U}) \times \R.
\end{equation}
Consider a class $\cal{H}_{\mb{a}}$ of nonlinear functions mapping from $\R^{2k} \times \cal{Q}$ to $\R^m$ and a class $\cal{H}_{b}$ of nonlinear functions mapping from $\R^{2k} \times \cal{Q} \times \R^n$ to $\R$. For a given $\widehat{\mb{a}} \in \cal{H}_{\mb{a}}$ and $\widehat{b} \in \cal{H}_b$, define $\widehat{\dot{W}}$ as:
\begin{equation}\label{eqn:estimator}
    \widehat{\dot{W}}(\bs{\eta}, \mb{q}, \dot{\mb{q}}, \mb{u}) = \widehat{\dot{V}}(\bs{\eta}, \mb{u}) + \widehat{\mb{a}}(\bs{\eta}, \mb{q})^\top\mb{u} + \widehat{b}(\bs{\eta}, \mb{q}, \dot{\mb{q}}),
\end{equation}
and let $\cal{H}$ be the class of all such estimators mapping $\R^{2k} \times \cal{Q} \times \R^n \times \cal{U}$ to $\R$. Defining a loss function $\cal{L}: \R \times \R \to \R_+$, the supervised regression task is then to find a function in $\cal{H}$ via empirical risk minimization (ERM):
\begin{equation}
\label{eqn:erm}
    \inf_{\substack{ \widehat{\mb{a}} \in \cal{H}_{\mb{a}} \\ \widehat{b} \in \cal{H}_{b}}} \frac{1}{N} \sum_{i=1}^N \cal{L}(\widehat{\dot{W}}(\bs{\eta}_i, \mb{q}_i, \dot{\mb{q}}_i, \mb{u}_i), \dot{V}_i).
\end{equation}
This experiment protocol can be executed either in simulation or directly on hardware. While being simple to implement, supervised learning critically assumes independently and identically distributed (i.i.d) training data. Each experiment violates this assumption, as the regression target of each data point is coupled with the input data of the next time step. As a consequence, standard supervised learning with sequential, non-i.i.d data collection often leads to error cascades \cite{le2016smooth}.

\section{INTEGRATING EPISODIC LEARNING \& CLFs}
\label{sec:learnclfs}
In this section we present the main contribution of this work: an episodic learning algorithm that captures the uncertainty present in the Lyapunov function derivative in a learned model and utilizes it in a quadratic program based controller.
% In this section we present the main contribution of this work in the form of an episodic learning algorithm that captures the uncertainty present in the Lyapunov function derivative in a learned model and utilizes it in a quadratic program based controller. To ensure the data for the learning problem is representative of the desired domain of operation, it is gathered via a sequence of experiments using controllers specified in a manner inspired by DAgger.

\subsection{Episodic Learning Framework}

Episodic learning refers to learning procedures that iteratively alternates between executing an intermediate controller (also known as a roll-out in reinforcement learning \cite{kober2013reinforcement}), collecting data from that roll-out, and designing a new controller using the newly collected data.
Our approach integrates learning $\mb{a}$ and $b$ with improving the performance and stability of the control policy $\mb{u}$ in such an iterative fashion. First, assume we are given a nominal state-feedback controller $\mb{u}: \cal{Q} \times \R^n \times \cal{I} \to \cal{U}$. With an estimator $\widehat{\dot{W}} \in \cal{H}$ as defined in (\ref{eqn:estimator}), we specify an augmenting controller as:
\begin{align}
\label{eqn:augcontroller}
\mb{u}'(\mb{q},\dot{\mb{q}},t) = &~\argmin_{\mb{u}'\in\R^m} J(\mb{u}')\nonumber \\
&\quad\mathrm{s.t. }~ \widehat{\dot{W}}(\bs{\eta}, \mb{q}, \dot{\mb{q}}, \mb{u}(\mb{q}, \dot{\mb{q}}, t) + \mb{u}')\leq -c_3\norm{\bs{\eta}}^2_2 \nonumber\\
&\quad\quad\quad \mb{u}(\mb{q}, \dot{\mb{q}}, t) + \mb{u}' \in \cal{U},
\end{align}
where $J: \R^m \to \R$ is any positive semi-definite quadratic cost function. 

\begin{algorithm}[t]
    \begin{algorithmic}
        \Require Control Lyapunov Function $V$, derivative estimate $\widehat{\dot{V}}_0$, model classes $\cal{H}_{\mb{a}}$ and $\cal{H}_b$, loss function $\cal{L}$, set of initial configurations $\cal{Q}_0$, nominal state-feedback controller $\mb{u}_0$, number of experiments $T$, sequence of trust coefficients $0 \leq w_1 \leq \cdots \leq w_T \leq 1$
        \Statex
        \State $D = \emptyset$ \Comment{Initialize data set}
        \For{$k = 1, \dots, T$}
            \State $(\mb{q}_0, \mb{0}) \leftarrow \textrm{sample}(\cal{Q}_0 \times \{\mb{0}\})$ \Comment{Get initial condition}
            \State $D_k \leftarrow \textrm{experiment}((\mb{q}_0, \mb{0}), \mb{u}_{k-1})$ \Comment{Run experiment}
            \State $D \leftarrow D \union D_k$ \Comment{Aggregate data set}
            \State $\widehat{\mb{a}}, \widehat{b} \leftarrow \textrm{ERM}(\cal{H}_{\mb{a}}, \cal{H}_b, \cal{L}, D, \widehat{\dot{V}}_0)$ \Comment{Fit estimators}
            \State $\widehat{\dot{V}}_k \leftarrow \widehat{\dot{V}}_0 + \widehat{\mb{a}}^\top\mb{u} + \widehat{b}$ \Comment{Update derivative estimator}
            \State $\mb{u}_{k} \leftarrow \mb{u}_0 + w_k \cdot \textrm{augment}(\mb{u}_0, \widehat{\dot{V}}_k)$ \Comment{Update controller}
        \EndFor
        \Statex
        \Return $\widehat{\dot{V}}_T, \mb{u}_T$
    \end{algorithmic}
    \caption{Dataset Aggregation for Control Lyapunov Functions (DaCLyF)}\label{alg}
\end{algorithm}

\begin{figure*}[htpb]
    \centering
    \includegraphics[scale=0.525]{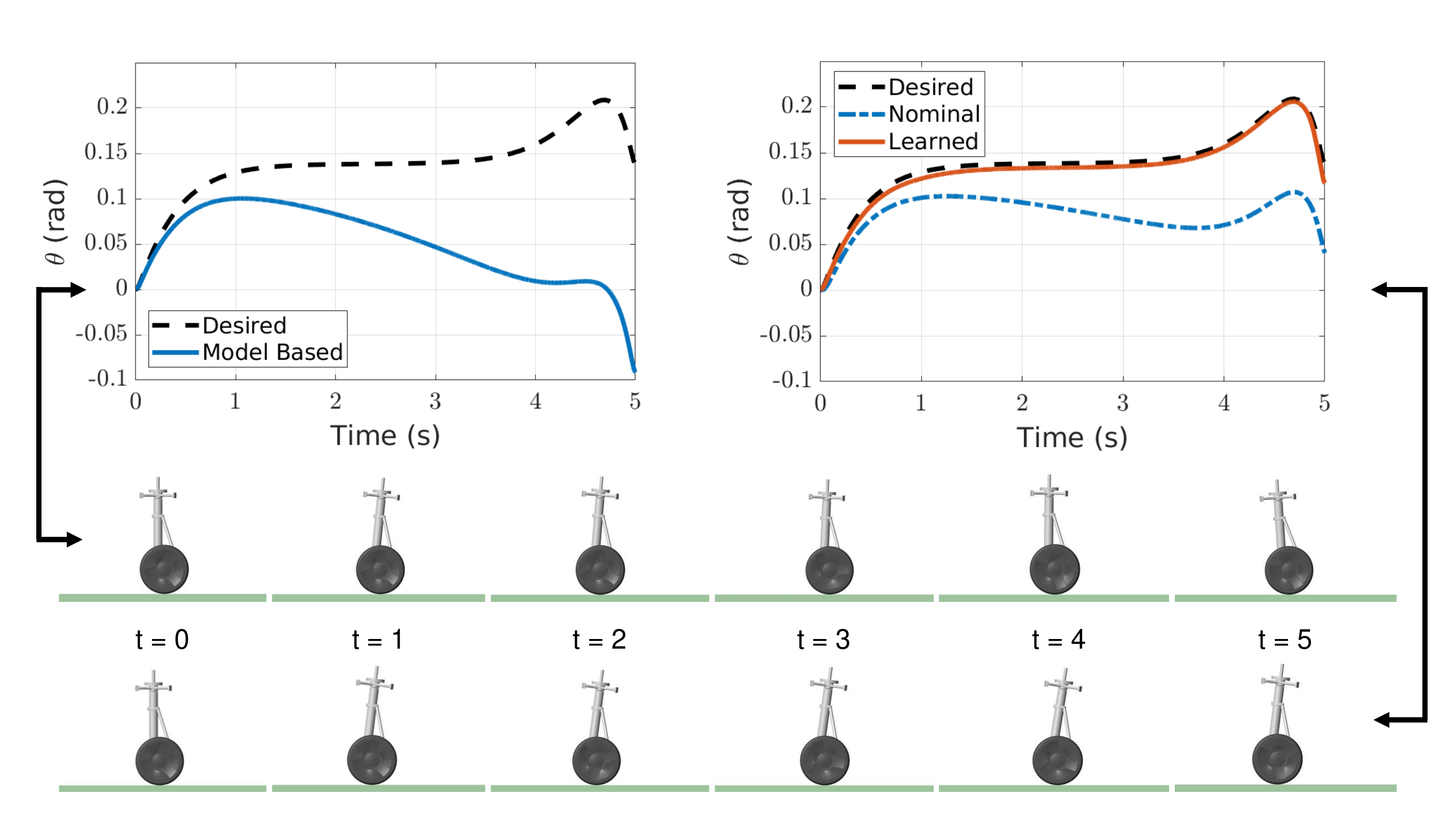}
    \caption{(Left) Model based QP controller fails to track trajectory. (Right) Improvement in angle tracking of system with augmented controller over nominal PD controller. (Bottom) Corresponding visualizations of state data. Note that Segway is tilted in the incorrect direction at the end of the QP controller simulation, but is correctly aligned during the augmented controller simulation.}
    \label{fig:trajectoryresults}
\end{figure*}

Our goal is to use this new controller to obtain better estimates of $\mb{a}$ and $b$. One option, as seen in Section \ref{sec:uncertainty_learning}, is to perform experiments and use conventional supervised regression to update $\widehat{\mb{a}}$ and $\widehat{b}$. To overcome the limitations of conventional supervised learning, we leverage reduction techniques: a sequential prediction problem is reduced to a sequence of supervised learning problems over multiple episodes \cite{ernst2005tree, ross2011reduction}. In particular, in each episode, an experiment generates data using a different controller. The data set is aggregated and a new ERM problem is solved after each episode. Our episodic learning implementation is inspired by the Data Aggregation algorithm (DAgger) \cite{ross2011reduction}, with some key differences:
\begin{itemize}
    \item DAgger is a reinforcement learning algorithm, which trains a policy directly in each episode using optimal computational oracles. Our algorithm defines a controller indirectly via a CLF to ensure stability.
    \item The ERM problem is underdetermined, i.e., different approximations $(\widehat{\mb{a}}, \widehat{b})$ may achieve similar loss for a given data set while failing to accurately model $\mb{a}$ and $b$. This potentially introduces error in estimating $\dot{V}$ for control inputs not reflected in the training data, and necessitates the use of exploratory control action to constrain the estimators $\widehat{\mb{a}}$ and $\widehat{b}$. Such exploration can be achieved by randomly perturbing the controller used in an experiment at each time step. This need for exploration is an analog to the notion of persistent excitation from adaptive systems \cite{narendra1987persistent}.
\end{itemize}

Algorithm \ref{alg} specifies a method of computing a sequence of Lyapunov function derivative estimates and augmenting controllers. During each episode, the augmenting controller associated with the estimate of the Lyapunov function derivative is scaled by a factor reflecting trust in the estimate and added to the nominal controller for use in the subsequent experiment. The trust coefficients form a monotonically non-decreasing sequence on the interval $[0, 1]$.
Importantly, this experiment need not take place in simulation; the same procedure may be executed directly on hardware. It may be infeasible to choose a specific configuration for an initial condition on a hardware platform; therefore we specify a set of initial configurations $\cal{Q}_0 \subseteq \cal{Q}$ from which an initial condition may be sampled, potentially randomly.

% While the algorithm is similar in structure to DAgger, key differences may be noted. First, while the DAgger algorithm trains a policy directly during each episode, our algorithm defines a controller indirectly by updating the Lyapunov function derivative estimate. Second, the assumptions required during analysis of DAgger are not made in our setting.

\subsection{Additional Controller Details}
During augmentation, we specify the controller in (\ref{eqn:augcontroller}) by selecting the minimum-norm cost function: 
\begin{equation}
    J(\mb{u}') = \frac{1}{2}\norm{ \mb{u}(\mb{q}, \dot{\mb{q}}, t) + \mb{u}' }_2^2,
\end{equation}
for all $\mb{u}' \in \R^m$, $\mb{q} \in \cal{Q}$, $\dot{\mb{q}} \in \R^n$, and $t \in \cal{I}$. We additionally incorporate a smoothing regularizer into the cost function of the form:
\begin{equation*}
    R(\mb{u}') = \overline{R} \norm{\mb{u}' - \mb{u}_{\mathrm{prev}}}_2^2,
\end{equation*} 
for all $\mb{u}' \in \R^m$, where $\mb{u}_{\mathrm{prev}} \in \R^m$ is the previously computed augmenting controller and $\overline{R} > 0$. This is done to avoid chatter that may arise from the optimization based nature of the CLF-QP formulation \cite{morris2013sufficient}.

Note that for this choice of Lyapunov function, the gradient $\derp{V}{\bs{\eta}}$, and therefore $\mb{a}$, approach $\mb{0}$ as $\bs{\eta}$ approaches $\mb{0}$, which occurs close to the desired trajectory. While the estimated Lyapunov function derivative may be fit with low absolute error on the data set, the relative error may still be high for states near the desired trajectory. Such relative error causes the optimization problem in (\ref{eqn:augcontroller}) to be poorly conditioned near the desired trajectory. We therefore add a slack term $\delta \in \R_+$ to the decision variables, which appears in the inequality constraint \cite{ames2013towards}. The slack term is additionally incorporated into the cost function as:
\begin{equation}
    C(\delta) = \frac{1}{2}\overline{C}\norm{ \left(\derp{V}{\bs{\eta}}\widehat{\mb{g}}(\mb{q})\right)^\top + \widehat{\mb{a}}(\bs{\eta}, \mb{q}) }_2^2\delta^2,
\end{equation}
for all $\delta \in \R_+$, where $\overline{C} > 0$. As states approach the trajectory, the coefficient of the quadratic term decreases and enables relaxation of the exponential stability inequality constraint. In practice this leads to input-to-state stable behavior, described in \cite{sontag1995characterizations}, around the trajectory.

The exploratory control during experiments is naively chosen as additive noise from a centered uniform distribution, with each coordinate drawn i.i.d. The variance is scaled by the norm of the underlying controller to introduce exploration while maintaining a high signal-to-noise ratio.
\begin{figure*}[tbph]
    \centering
    \includegraphics[scale=0.25, clip, trim={0in, 2.57in, 0.75in, 2.57in}]{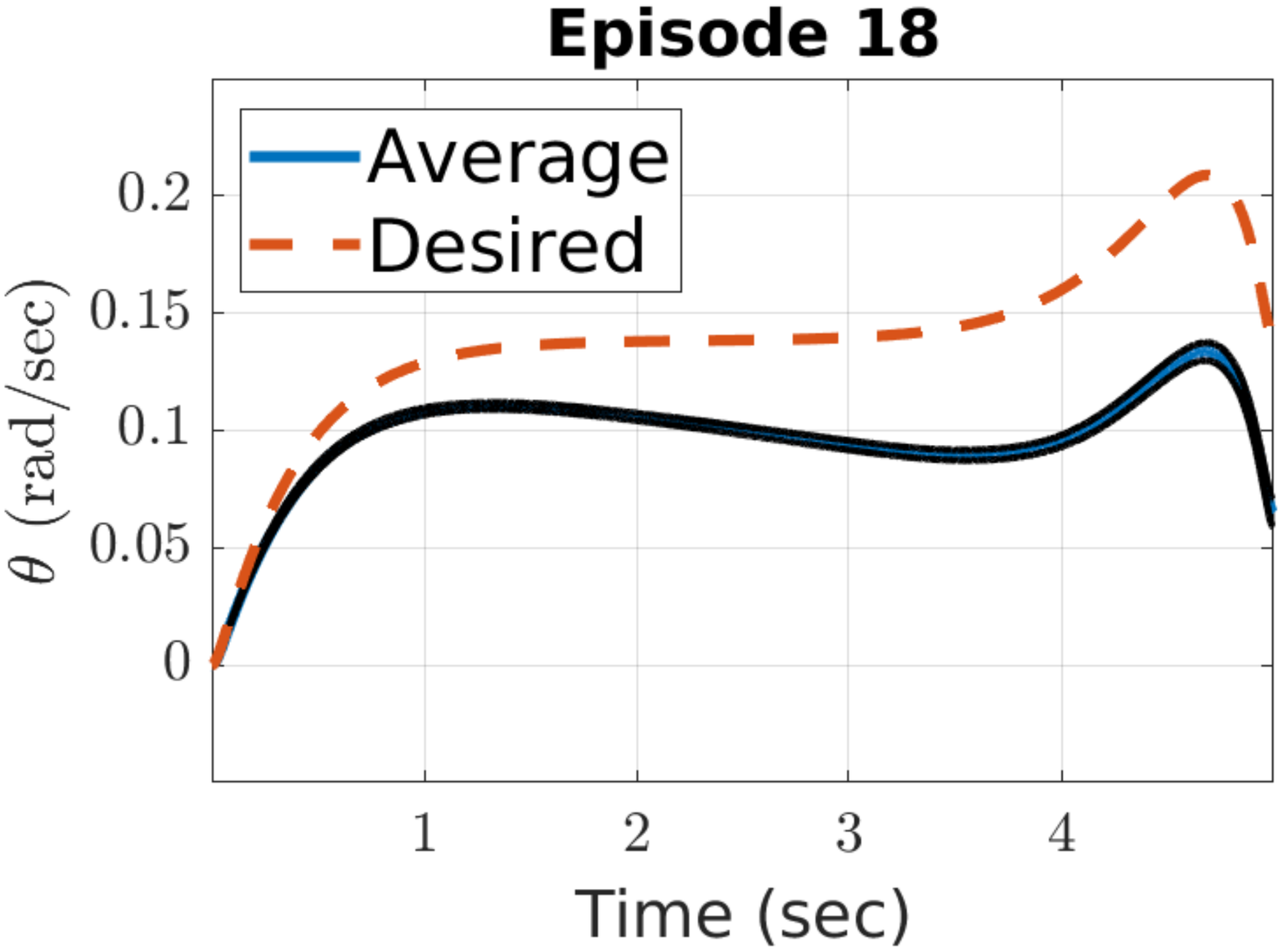}
    \includegraphics[scale=0.25, clip, trim={1.35in, 2.57in, 0.75in, 2.57in}]{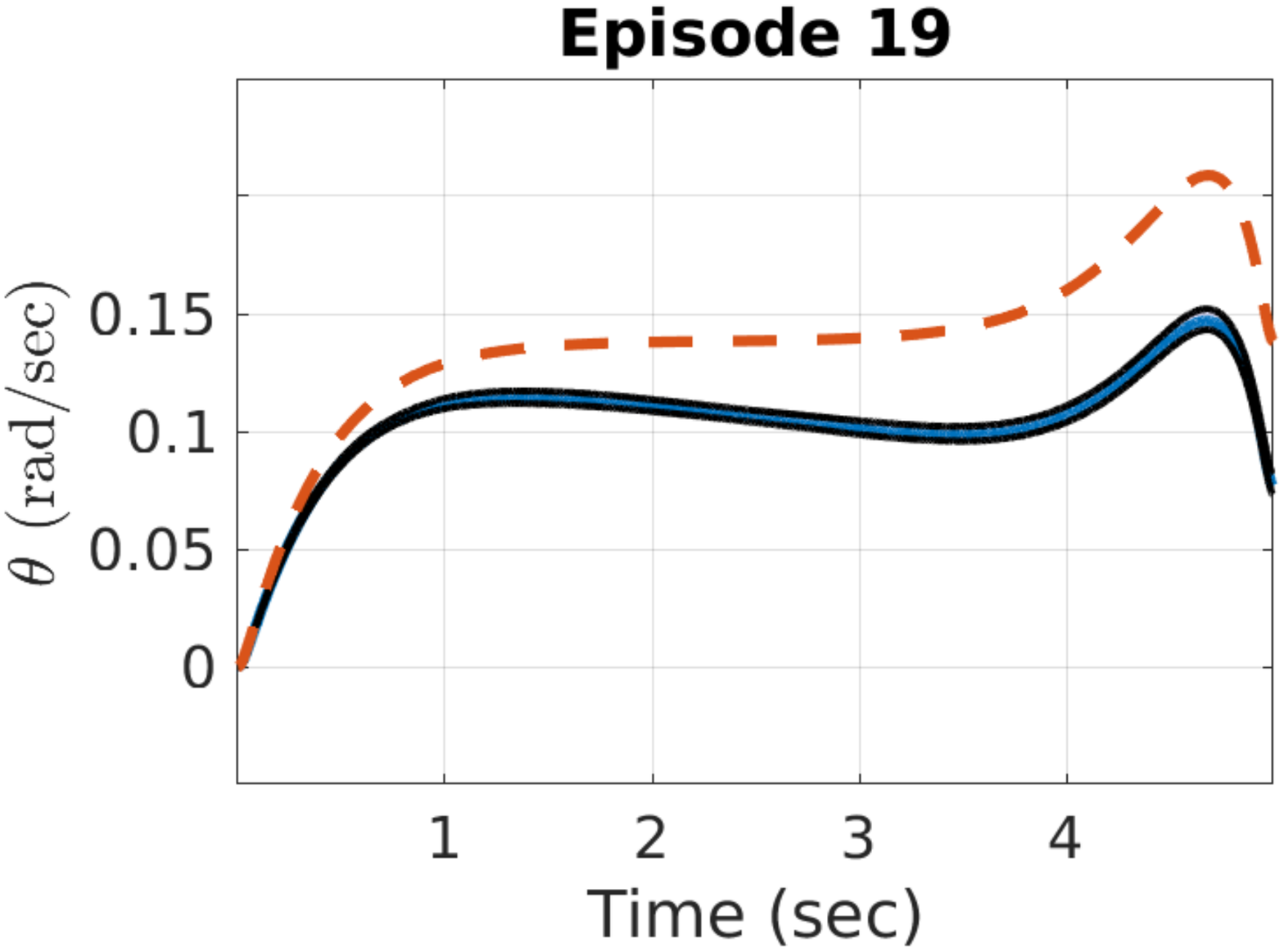}
    \includegraphics[scale=0.25, clip, trim={1.35in, 2.57in, 0.75in, 2.57in}]{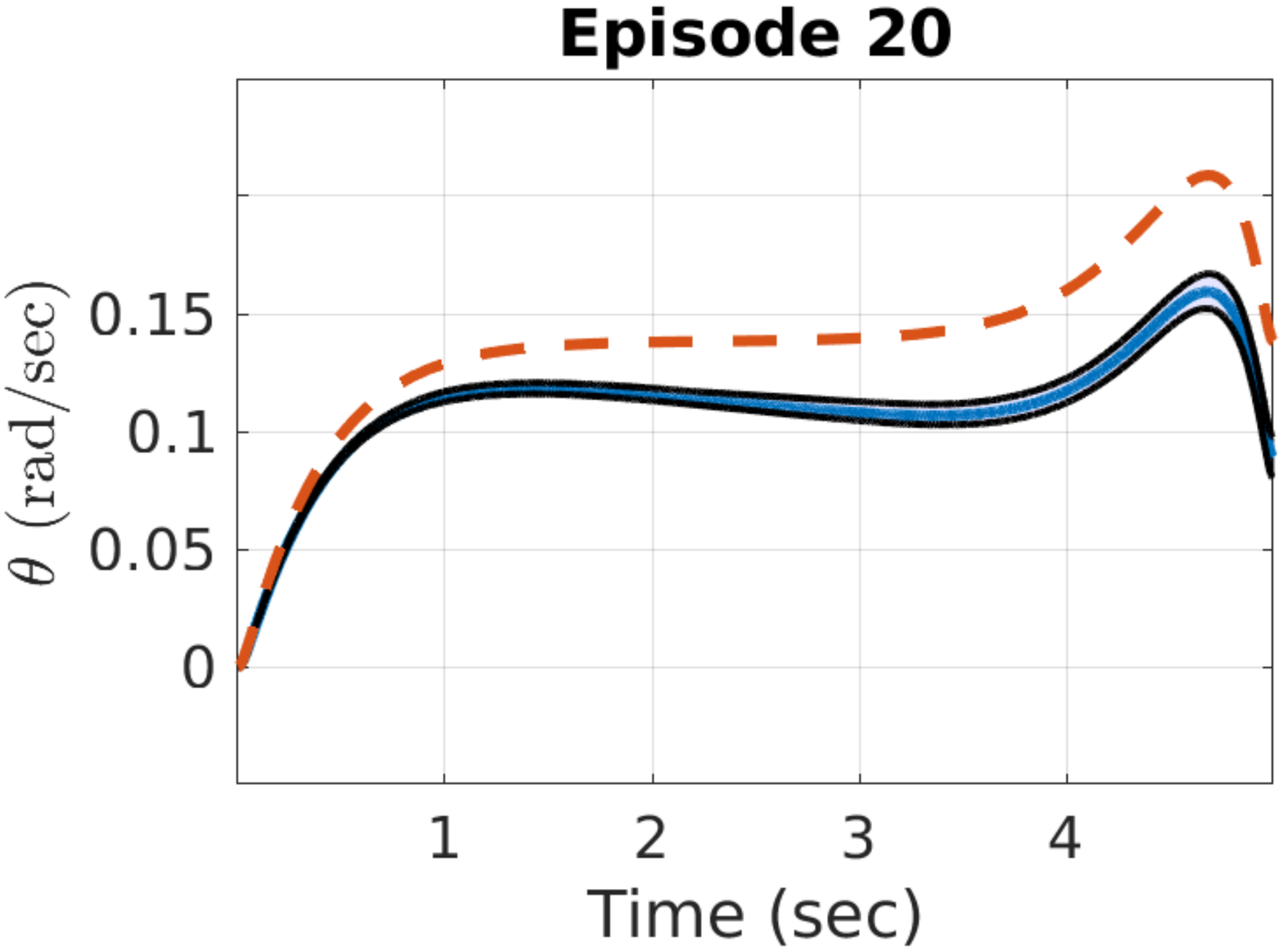}
    \includegraphics[scale=0.25, clip, trim={1.35in, 2.57in, 0.75in, 2.57in}]{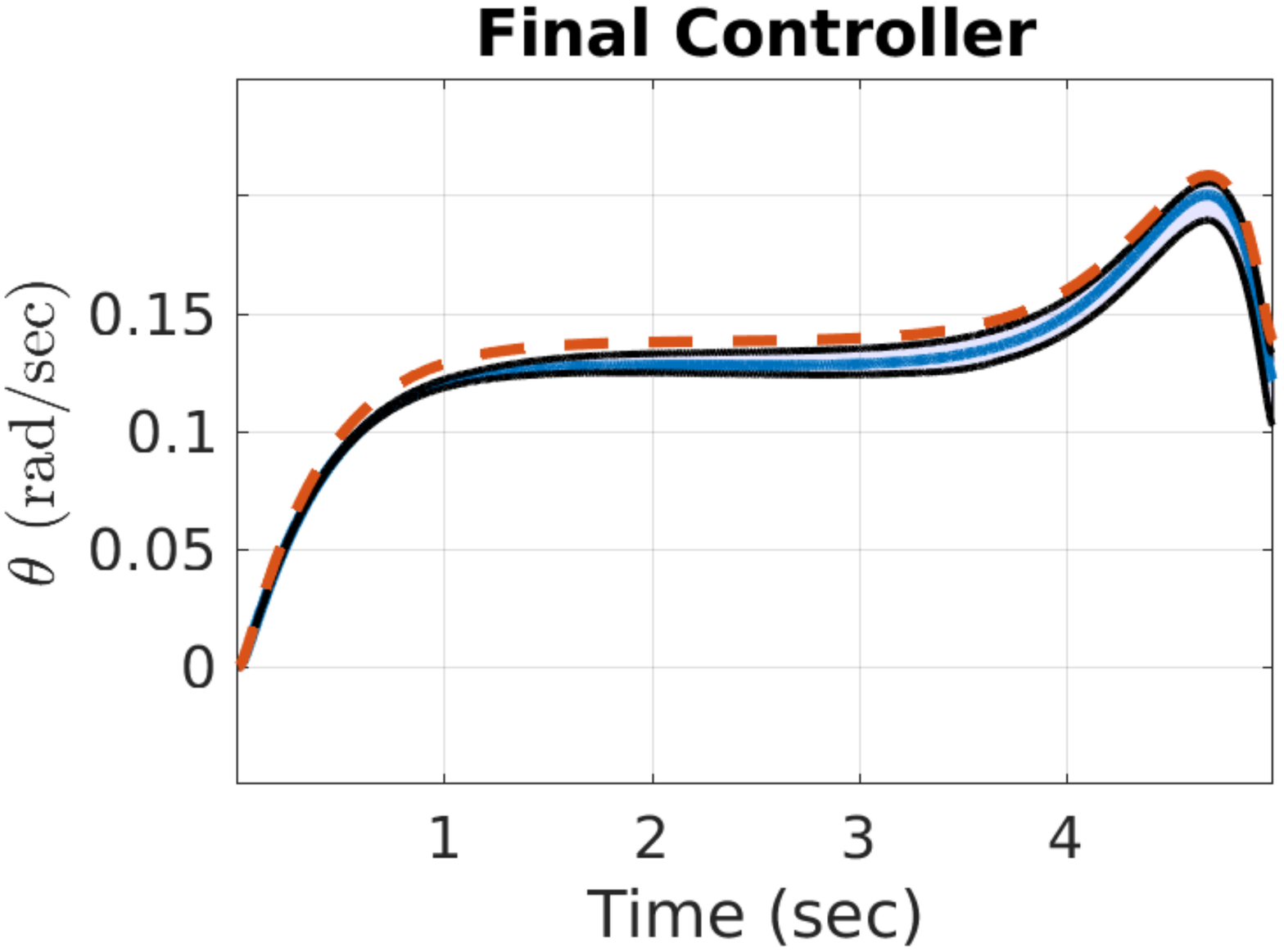}
    \caption{Augmenting controllers consistently improve trajectory tracking across episodes. 10 instances of the algorithm are executed with the shaded region formed from minimum and maximum angles for each time step within an episode. The corresponding average angle trajectories are also displayed.}
    \label{fig:augeps}
\end{figure*}
\section{APPLICATION ON SEGWAY PLATFORM}
\label{sec:simulation}

In this section we apply the episodic learning algorithm constructed in Section \ref{sec:learnclfs} to the Segway platform. In particular, we consider a 4-dimensional planar Segway model based on the simulation model in \cite{gurriet2018towards}. The system states consist of horizontal position and velocity, pitch angle, and pitch angle rate. Control is specified as a single voltage input supplied to both motors. The parameters of the model (including mass, inertias, and motor parameters but excluding gravity) are randomly modified by up to 10\% of their nominal values and are fixed for the simulations.

We seek to track a pitch angle trajectory\footnote{Trajectory was generated using the GPOPS-II Optimal Control Software} generated for the estimated model. The nominal controller is a linear proportional-derivative (PD) controller on angle and angle rate error. 20 experiments are conducted with trust values varying from 0.01 to 0.99 in a sigmoid fashion. The exploratory control is drawn uniformly at random between $-20\%$ and $20\%$ of the norm of the underlying controller for an episode for the first 10 episodes. The percentages decay linearly to 0 in the remaining 10 episodes. The model classes selected are sets of two-layer neural networks with ReLU nonlinearities with hidden layer width of 2000 nodes\footnote{Models were implemented in Keras}. The inputs to both models are all states and the Lyapunov function gradient.

Failure of the controller (\ref{eqn:qpcontroller}) designed for the estimated model to track the desired trajectory is seen in the left portion of Fig. \ref{fig:trajectoryresults}. The baseline PD controller and the augmented controller after 20 experiments can be seen in the right portion Fig. \ref{fig:trajectoryresults}. Corresponding visualizations of the Segway states are displayed at the bottom of Fig. \ref{fig:trajectoryresults}. The augmented controller exhibits a notable improvement over the model-based and PD controller in tracking the trajectory.   

To verify the robustness of the learning algorithm, the 20 experiment process was conducted 10 times. After each experiment the intermediate augmented controller was tested without exploratory perturbations. For the last three experiments and a test of the final augmented controller, the minimum, mean, and maximum angles across all 10 instances are displayed for each time step in Fig. \ref{fig:augeps}. The mean trajectory consistently improves in these later episodes as the trust factor increases. The variation increases but remains small, indicating that the learning problem is robust to randomness in the initialization of the neural networks, in the network training algorithm, and in the noise added during the experiments. The performance of the controller in the earlier episodes displayed negligible variation from the baseline PD controller due to small trust factors.

LyaPy is available at \href{https://github.com/vdorobantu/lyapy}{https://github.com/vdorobantu/lyapy}.

\section{CONCLUSIONS \& FUTURE WORK}
We presented an episodic learning framework that directly integrates into an established method of nonlinear control using CLFs. Our method allows for the effects of both parametric error and unmodeled dynamics to be learned from experimental data and incorporated into an quadratic program controller. The success of this approach was demonstrated in simulation on a Segway, showing improvement upon a model estimate based controller. 

There are two main interesting directions for future work. First, a more thorough investigation of episodic learning algorithms can yield superior performance as well as learning-theoretic converge guarantees.  Other episodic learning approaches to consider include SEARN \cite{daume2009search}, AggreVaTeD \cite{sun2017deeply}, and MoBIL \cite{cheng2019accelerating}, amongst others.
%stability guarantees based on collected data sets, determining optimal but stable exploration policies, analyzing performance on more complex multiple-input systems, integrating this algorithm formally with the DAgger framework,
Second, our approach can also be applied to learning with other forms of guarantees, such as with Control Barrier Functions (CBFs) \cite{ames2017control}.  Existing work on learning CBFs are restricted to learning with Gaussian processes \cite{wang2018safe,fisac2018general,cheng2019aaai}, and also learn over the full state space rather than over the low-dimensional projection onto the CBF time derivative. 
%exploring applications to Control Barrier Functions (CBFs) 
 %, and demonstrating this approach on a physical Segway platform.

% \begin{figure}
%     \centering
%     \includegraphics[scale=0.54]{qp.eps}
%     \caption{Model based QP controller fails to track trajectory.}
%     \label{fig:qpfailure}
% \end{figure}

% \begin{figure}[thpb]
%     \centering
%     \includegraphics[scale=0.55]{pd_aug_comp.eps}
%     \caption{Improvement in angle tracking of system with augmenting controller over nominal controller (PD).}
%     \label{fig:pdaugcomp}
% \end{figure}

\addtolength{\textheight}{-1.5cm}   % This command serves to balance the column lengths

\bibliographystyle{plain}
\bibliography{main}

\end{document}